\documentclass[runningheads]{llncs}
\usepackage{graphicx}
\usepackage{amsmath,amssymb} 
\usepackage{color}
\usepackage{multirow}
\usepackage[breaklinks=true,bookmarks=false]{hyperref}

\usepackage[normalem]{ulem}
\useunder{\uline}{\ul}{}
\begin{document}
\title{Efficient Semantic Scene Completion Network with Spatial Group Convolution}

\titlerunning{Semantic Scene Completion with Spatial Group Convolution}
%
%
\authorrunning{Jiahui Zhang, Hao Zhao and et al.}
%

\author{Jiahui Zhang\thanks{indicates interns at Intel Labs China. $^\boxtimes$ indicates corresponding authors.} \inst{1}
\and
Hao Zhao${^\star}$ \inst{2} \and
Anbang Yao$^{\boxtimes}$\inst{3} \and
Yurong Chen\inst{3} \and
Li Zhang\inst{2} \and
Hongen Liao$^{\boxtimes}$\inst{1}
}

\institute{Department of Biomedical Engineering, Tsinghua University \and
Department of Electronic Engineering, Tsinghua University \and
Intel Labs China \\
\email{\{jiahui-z15@mails.,zhao-h13@mails.,chinazhangli@mail.,liao@\}tsinghua.edu.cn \\
\{anbang.yao,yurong.chen\}@intel.com}}

\maketitle              
\begin{abstract}
We introduce Spatial Group Convolution (SGC) for accelerating the computation of 3D dense prediction tasks. SGC is orthogonal to group convolution, which works on spatial dimensions rather than feature channel dimension. It divides input voxels into different groups, then conducts 3D sparse convolution on these separated groups. As only valid voxels are considered when performing convolution, computation can be significantly reduced with a slight loss of accuracy. The proposed operations are validated on semantic scene completion task, which aims to predict a complete 3D volume with semantic labels from a single depth image. With SGC, we further present an efficient 3D sparse convolutional network, which harnesses a multiscale architecture and a coarse-to-fine prediction strategy. Evaluations are conducted on the SUNCG dataset, achieving state-of-the-art performance and fast speed. Code is available at \url{https://github.com/zjhthu/SGC-Release.git}
\keywords{Spatial Group Convolution, Sparse Convolutional Network, Efficient Neural Network, Semantic Scene Completion}
\end{abstract}

\section{Introduction}
\label{sec::intro}
3D shape processing has attracted increasing attention recently, because large scale 3D datasets and deep learning based methods open new opportunities for understanding and synthesizing 3D data, such as segmentation and shape completion.
These 3D dense prediction tasks are quite useful for many applications. For example, robots need semantic information to understand the world, while knowing complete scene geometry can help them to grasp objects \cite{2016arXiv160908546V} and avoid obstacles.
However, it is not a trivial task to adopt 3D Convolutional Neural Network (CNN) by just adding one dimension to 2D CNN. Dense 3D CNN methods \cite{Wu2015,Maturana2015} face the problem of cubic growth of computational and memory requirements with the increase of voxel resolution.

\begin{figure}[t]
\centering
\includegraphics[width=\textwidth]{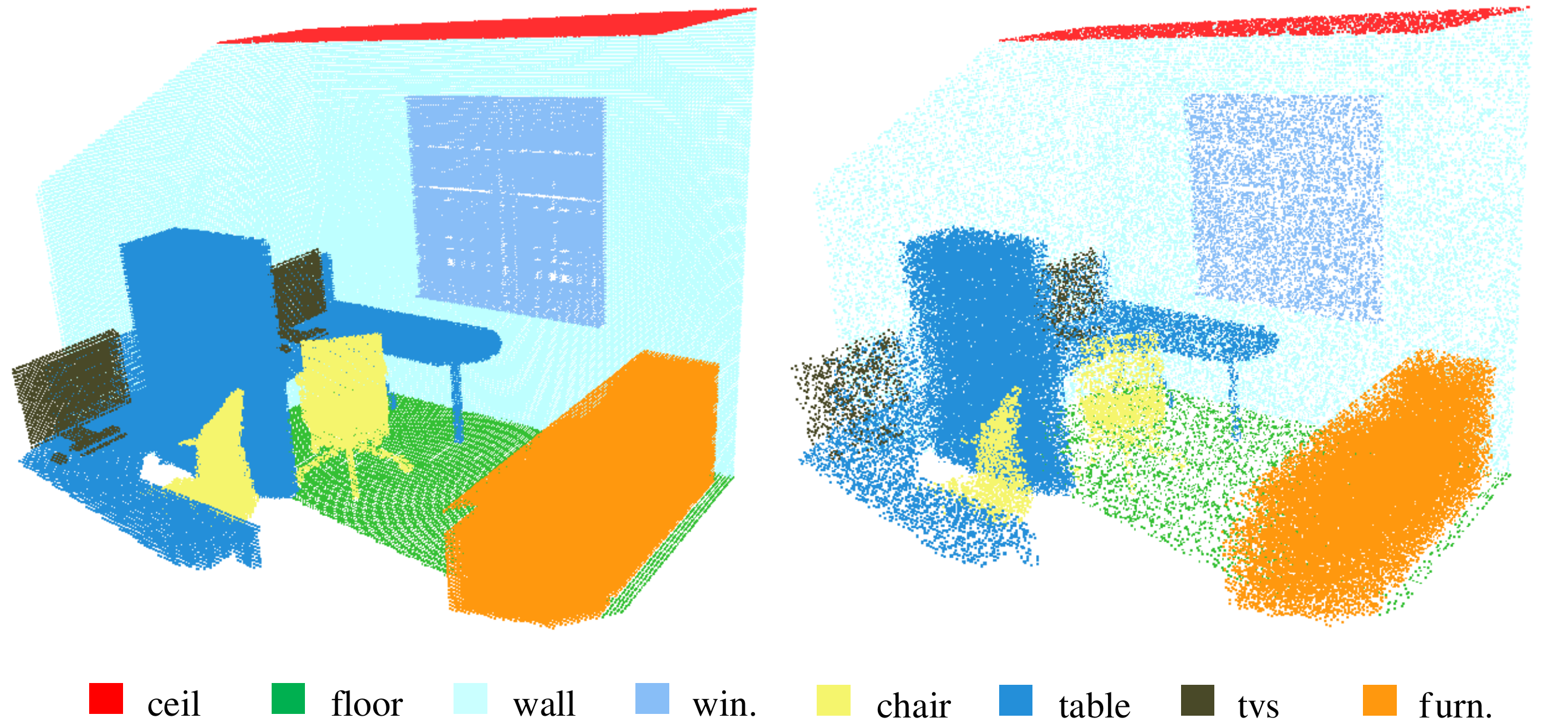}
\caption{A 3D scene image from the SUNCG dataset. Left is the ground truth image. Right is a sampled image with only 30\% voxels reserved. Giving only partial voxels does not prevent humans in reasoning the overall semantic information, but it imposes a challenge to recognize small objects such as chair{'}s leg. (\textbf{Best viewed in color})}
\label{fig:full_sparse}
\end{figure}

But meanwhile, we observe that 3D data has some attractive characteristics, which inspire us to build efficient 3D CNN blocks. Firstly, intrinsic sparsity in 3D data. Most of the voxels in a dense 3D grid are empty. Non-trivial voxels usually exist near the boundaries of objects. This property has been explored in several recent works \cite{Riegler2016,wang2017cnn,Graham2015,2017arXiv170601307G}.
Secondly, redundancy in 3D voxels. Dense 3D voxels are usually redundant, discarding a large portion of voxels (e.g. 70\%) randomly does not prevent humans from reasoning the overall semantic information, as shown in Fig. \ref{fig:full_sparse}.
Thirdly, different subsets of original dense voxels contain complementary information. It is hard to recognize objects with small size and complex geometry when giving only partial voxels.
These properties motivate us to design computation-efficient 3D CNNs for dense prediction tasks.
We adopt Sparse Convolutional Network (SCN) \cite{2017arXiv170601307G}\footnote{or called Submaniflod Sparse Convolutional Network in \cite{2017arXiv170601307G}.} to exploit the intrinsic sparsity of 3D data, which encodes sparse 3D data with Hash Table and presents sparse convolution design. These designs can avoid unnecessary memory or computation cost on empty voxels.
However, the computation is still intensive when the resolution is high or input is not so sparse.
For example, the complexity of the baseline SCN used in this paper is about 80 GFLOPs while only outputting $1/64$ sized predictions.
Our work takes advantage of SCN and steps further by encouraging higher sparsity in feature maps.
We propose SGC to exploit the redundancy of 3D voxels, which partitions features into different groups and makes voxels sparser. Then we conduct sparse convolution on each group.
Because only valid voxels are considered in sparse convolution rather than all voxels in a regular grid, and only partial voxels exist in each group after partition, the computation of networks with SGC can be significantly reduced compared to previous SCN.
Besides, in order to utilize the complementary information of different groups, results of different groups after certain SGC operations are gathered for further processing.

Network acceleration methods in 2D CNN such as weight pruning, quantization, and Group Convolution (GC) design \cite{2017arXiv170404861H,zhang2018shufflenet} can also be used, but these methods have not been well explored in 3D CNNs for now. Different from these methods, \textbf{SGC speeds up 3D CNNs from another perspective by encouraging sparsity in feature maps}.
Though recently there are works \cite{engelcke2017vote3deep,hackel2018inference} exploiting sparsity in feature maps, they are not suitable for dense prediction tasks
because some voxels need to be predicted are deactivated in the network.
\textbf{Our method is orthogonal to Group Convolution}, which is an operation widely used in recent CNN architectures \cite{krizhevsky2012imagenet,xie2017aggregated,chollet2017xception}. SGC is defined on spatial dimensions while GC is defined on channel dimension. Besides, because voxels in different groups are similar, weights are shared between different groups in SGC, which is not the case in GC.

We validate our method on semantic scene completion as test case to show its effectiveness on 3D dense prediction tasks. This task not only aims to predict semantic labels, but also needs to output complete structure which is different from the input.
We introduce a novel SCN architecture that is applicable to scenarios where output has a different structure with input. Dense deconvolution layer and Abstracting Module are designed to generate voxels which are absent in input and remove trivial voxels respectively. Multiscale encoder-decoder architecture and coarse-to-fine prediction strategy are used for final predictions.
We evaluate our network on the SUNCG dataset \cite{song2016semantic} and achieve state-of-the-art results.
Our SGC operation can reduce about $3/4$ of the computation while losing only 0.7\% and 1.2\% in terms of Intersection over Union (IoU) for scene completion and semantic scene completion compared to networks without SGC.

Our main contributions are as follows:
\begin{itemize}
  \item We propose SGC by exploiting sparsity in features for 3D dense prediction tasks, which can significantly reduce computation with slight loss of accuracy.
  \item We present a novel end-to-end sparse convolutional network design to generate unknown structures for 3D semantic scene completion.
  \item We achieve state-of-the-art results on the SUNCG dataset, reaching an IoU of 84.5\% for scene completion and 70.5\% for semantic scene completion.
\end{itemize}

\section{Related works}

\subsection{3D Deep Learning}
\label{sec:3DDL}
The success of deep learning in 2D computer vision areas has inspired researchers to  employ CNN in 3D tasks, such as object recognition \cite{Wu2015,Maturana2015,Qi}, shape completion \cite{Wu2015,song2016semantic,Dai2016}, and segmentation \cite{song2016semantic,Ben-Shabat2017}. However, the cubic growth in data size impedes building wider and deeper networks because of memory and computation restrictions. Recently, several works attempt to solve this problem by utilizing the intrinsic sparsity of 3D data. FPNN \cite{Li2016a} used learned field probes to sample 3D data at a small set of positions, then fed features into fully connected layers. Graham et al. \cite{Graham2015,2017arXiv170601307G} proposed Hash Table based sparse convolutional networks and solved the ``submanifold dilation" problem by forcing to keep the same sparsity level throughout the network.
OctNet \cite{Riegler2016} and O-CNN \cite{wang2017cnn} used Octree-based 3D CNN for 3D shape analysis. SBNet \cite{ren2018sbnet} performed convolution on blockwise decomposition of the structured sparsity patterns. Apart from these methods based on volumetric representation, PointNet \cite{qi2017pointnet} is a seminal work building deep neural networks directly on point clouds. PointNet++ \cite{qi2017pointnet++} and Kd-Networks \cite{klokov2017escape} further employed hierarchical architectures to capture local structures of point clouds.

Our main difference with these architectures is the introduction of SGC, which encourages higher sparsity in features and makes networks more efficient.

\subsection{Computation-efficient Networks}
Most previous computation-efficient networks focus on reducing model size to accelerate inference, such as pruning weight connections \cite{LeCun:1989:OBD,han2015learning} and quantizing weights \cite{gong2014compressing}.
Another line of works uses GC to reduce the computation, such as MobileNet \cite{2017arXiv170404861H} and ShuffleNets \cite{zhang2018shufflenet}. GC separates features to different groups along channel dimension and performs convolution on each group parallelly. Besides, Graham \cite{Graham2015} used smaller filters on different lattices to decrease the computation.

However, there are seldom works designing computation-efficient networks by exploiting higher sparsity in feature maps for 3D dense prediction tasks. Vote3deep \cite{engelcke2017vote3deep} encouraged sparsity in feature maps using $L_1$ regularization. ILA-SCNN \cite{hackel2018inference} used adaptive rectified linear unit to control the sparsity of features. But these methods are not suitable for dense prediction tasks, because some desired voxels are deactivated in the network and cannot be recovered. Besides, Li et al. \cite{2017arXiv170401344L} also exploited sparsity and reduced the computation of 2D segmentation task with cascaded networks, and only hard pixels are handled by deeper sub-models.

Different with these methods, we create groups along the spatial dimensions and make voxels in each group sparser. Computation of convolution can be largely reduced because only partial valid voxels are used in each computation.

\subsection{3D Semantic Segmentation and Shape Completion}
3D semantic segmentation \cite{Yi2017,chang2017matterport3d,Liu2017,Qi2017} and Shape Completion \cite{Wu2015,Firman2016,Dai2016,Han2017} are both active areas in computer vision. 3D segmentation gives semantic labels to observed voxels, while shape completion completes missing voxels. SSCNet \cite{song2016semantic} combined these two tasks together and showed that segmentation and completion can benefit from each other. In order to generate high resolution 3D structure, various methods had been explored, such as long short-term memorized \cite{Han2017}, coarse-to-fine strategy \cite{dai2018scancomplete}, 3D generative adversarial network \cite{yang20183d}, and inverse discrete cosine transform \cite{johnston2017scaling}.
Recently, segmentation and completion are both benefited from these advanced 3D deep learning methods described in section \ref{sec:3DDL}. Different methods have been presented in the 3D segmentation challenge \cite{Yi2017ShapeNet}, such as SCN, Pd-Network, densely connected PointNet, and Point CNN \cite{2018arXiv180107791L}. For 3D completion tasks, advanced Octree-based CNN methods \cite{Riegler2017,Tatarchenko2017,Hane2017} were also used for generating high resolution 3D outputs. Our network architecture shares some similarities with \cite{Riegler2017,Tatarchenko2017},
while the main difference is that we focus on efficient model design in this paper.

\section{Method}
In this section, we firstly give a brief introduction to previous SCN architecture \cite{2017arXiv170601307G}, and then introduce SGC for computation-efficient 3D dense prediction tasks. Thirdly, a novel sparse convolutional network architecture which can predict unknown structures will be presented for semantic scene completion. Finally, details about training and networks will be given.

\subsection{Sparse Convolutional Network}
\label{sec::sparse_conv}
Previous dense 3D convolution is neither computational nor memory efficient because of the usage of dense 3D grid for representation.
Another problem is that traditional ``dense" convolution has the ``dilation" problem \cite{2017arXiv170601307G} which will destroy the sparsity of 3D feature maps.
For example, after a $3\times3\times3$ convolution, surrounding 26 voxels will be filled in.
SCN addressed these problems by only storing non-empty voxels in 3D feature maps using Hash Table. \textbf{Only non-empty voxels are considered in sparse convolutional network}. Besides, \textbf{it forces to keep sparsity at the same level throughout the network} when performing convolution, which means the activation pattern of next layer is the same as the previous layer. These designs can largely decrease computation and memory requirements, enabling the usage of deeper 3D CNNs.

However, there is still intensive computation in 3D sparse CNN as mentioned above. Thus reducing the computation of 3D sparse CNN is necessary for real-time applications. Another problem of previous SCN is that it cannot be directly used for scene completion task. Because completion needs to output a complete structure which is different from the input, while previous SCN can only output predictions with the same structure as input. We introduce a novel sparse convolutional network to predict unknown structures.
\begin{figure}[t]
\centering
\includegraphics[width=\textwidth]{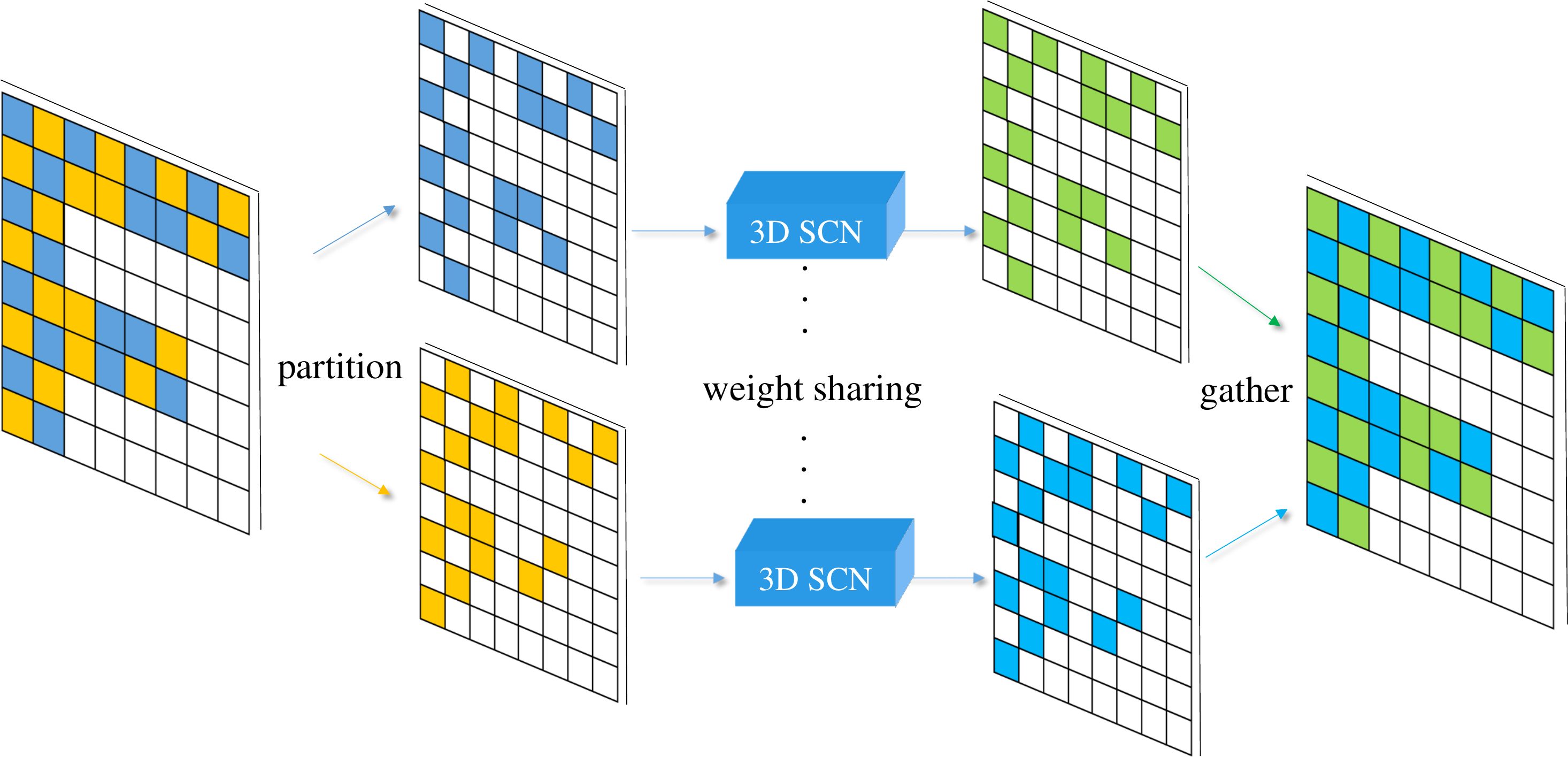}
\caption{Illustration of SGC. Feature maps are partitioned uniformly into different groups along the spatial dimensions (only two groups are shown here). 3D CNNs are conducted on different groups and give the final dense prediction for all voxels. Weights are shared between different groups.}
\label{fig:spatialGroupConv}
\end{figure}

\subsection{Spatial Group Convolution}
This section introduce SGC which can significantly reduce the computation of 3D dense prediction tasks.
Our design makes use of those three properties of 3D data described in section \ref{sec::intro} (see Fig. \ref{fig:spatialGroupConv}). We partition voxels uniformly into different groups, then conduct 3D sparse convolution on each group.
Weights are shared among different groups because these groups are similar. Features of different groups are gathered later in order to utilize the complementary information of different groups.

In the implementation of SGC, we partition features along the spatial dimensions and then stack different groups along the batch dimension. For one sparse feature map whose size is $B\times D \times H \times W \times C$ ($batch size \times depth \times height \times width \times channel$), after the partition operation, it becomes $(G \times B)\times D \times H \times W \times C$, where $G$ is the group number. Note that because we use Hash Table based representation, only non-empty voxels are stored. So this operation does not require extra memory. In each convolution computation, only part of original non-empty voxels in its receptive filed participate in the calculation, and the number of valid voxels in each group is about $1/G$ of the original non-empty voxels after partition. The final computation cost is thus about $1/G=\frac{N \times \frac{1}{G}\times k^3}{N \times k^3}$ of original convolution when ignoring the bias computation, where $N$ is the total number of valid voxels, and $k$ is the filter size. SGC can readily replace plain 3D sparse convolution in existing CNNs.


Obviously, partition strategy plays an important role in SGC. Here we present two different partition strategies:
\begin{itemize}
  \item Random partition method. Voxels of feature maps are partitioned into different groups randomly and uniformly.
  \item Partition with a fixed pattern. Random partition expects convolutional filters to be invariant to all possible patterns of activation, which may be hard for CNN to learn. We propose to partition input voxels with a fixed pattern for all input voxels throughout training and testing. For example, we can partition voxels by the following formulation:

\begin{equation}
\label{equ:pattern}
    i = mod(ax+by+cz,G)
\end{equation}

where $i$ is the group index, $(x,y,z)$ is the position of the voxel, $G$ is the total group number, $mod$ is the modulus operation, $(a,b,c)$ controls the distribution of different groups. This strategy can also partition voxels uniformly but in a fixed pattern manner. Different $(a,b,c)$ and $G$ give different patterns.
\end{itemize}

\subsection{Sparse Convolutional Networks for Semantic Scene Completion}
This section will present a novel SCN architecture for semantic scene completion. Previous SCN \cite{GrahamSCN} keeps the sparsity unchanged to avoid ``submanifold dilation" problem. The output of previous SCN has the same known structure as input. This design restricts its application in shape completion, RGB-D fusion and etc., which aim to predict unknown structures. In order to generate unknown voxels for semantic scene completion task, we have to break this restriction.

\begin{figure}[t]
\centering
\includegraphics[width=\textwidth]{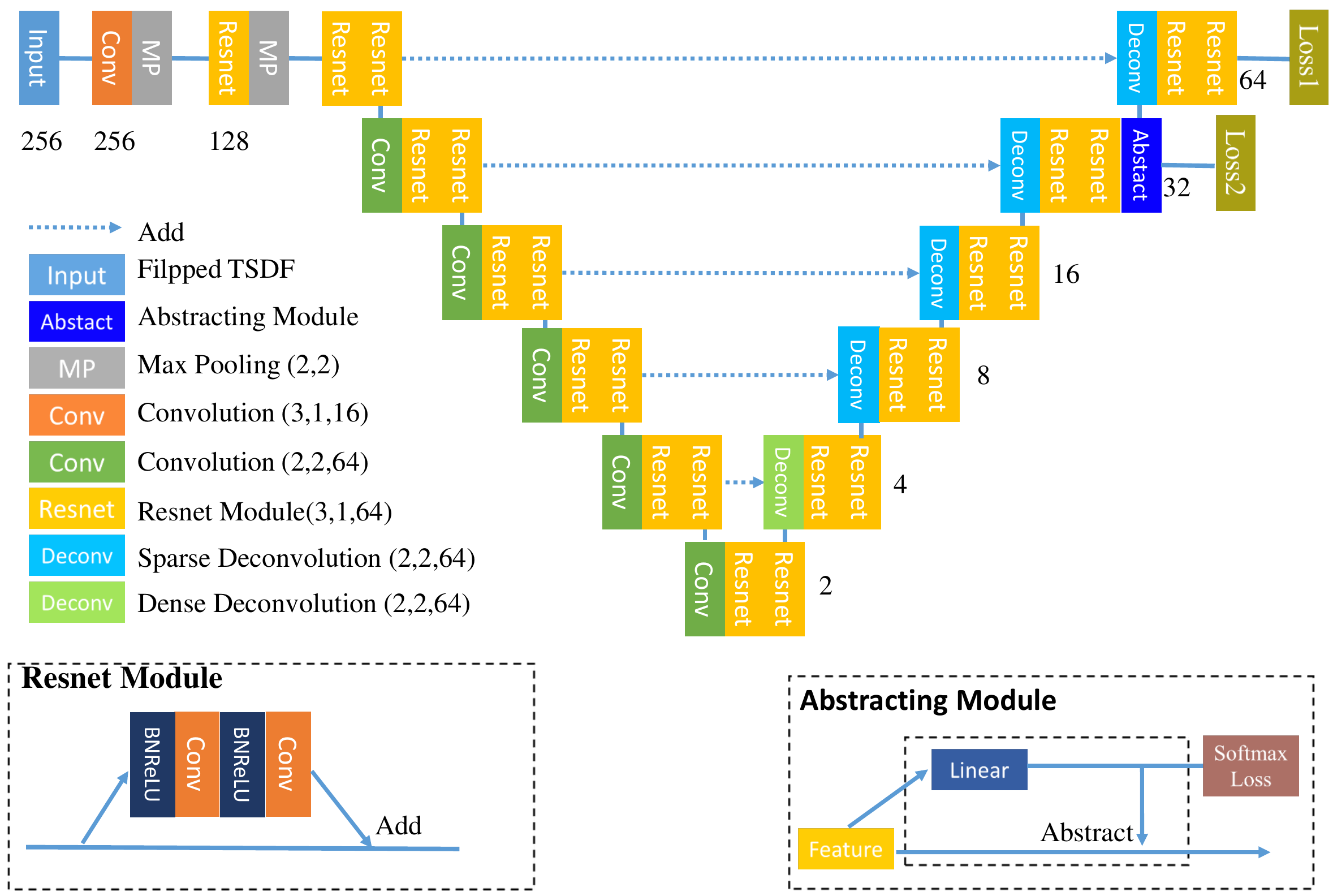}
\caption{Network architecture for semantic scene completion. Taking flipped TSDF as input, the network predicts occupancy and object labels in $1/4$ size. \textbf{The resolution of each layer is marked nearby.} Parameters of each layer are shown in the order of (filter size, stride, output channel). Dense deconvolution layers can generate new voxels. The Abstracting module can abstract non-trivial voxels to high resolution according to the prediction in low resolution. (\textbf{Best viewed in color})}
\label{fig:network}
\end{figure}

Here we use multiscale encoder-decoder architecuture \cite{ronneberger2015u}.
As shown in Fig. \ref{fig:network}, encoder modules are constituted of sparse convolutions described in section \ref{sec::sparse_conv}. While in decoder modules, we implement a ``dense" deconvolution layer to generate new voxels.
More specifically, after a ``dense" up-sampling deconvolution  layer, each voxel in low resolution will generate $2\times2\times2$ voxels in high resolution. The sparsity changes rather than keeping the same as the layers in encoder modules.
New voxels can be generated in this process.

Applying this module repeatedly in each scale can generate all missing structures but it will soon destroy the sparsity of 3D feature maps just as the ``submanifold dilation" problem.
So we introduce Abstracting Module similar to \cite{Riegler2017,Tatarchenko2017}, which abstracts a coarse structure and removes unnecessary voxels in low resolution. Details will be refined in high resolution.
The Abstracting Module contains a $1\times1\times1$ convolution layer and a softmax layer, these layers give a prediction in this scale and provide guiding information for abstracting. Only voxels with non-empty labels and their surrounding voxels are abstracted.
Abstracting these surrounding empty voxels within a distance of $k$ could provide fine details. $k=1$ works well in our practice.
We apply the Abstracting module in resolution higher than 32 because removing voxels in early stages may hurt the performance. Since our setting exploits resolution 64 for output, one Abastract module is enough.

Voxel-wise softmax loss is used in the two scales which give a prediction:
\begin{equation}
\label{equ:w_softmax}
    L_i={\frac{1}{\sum{w_j}} \sum\limits_j{w_j L_{sm}(p_j, y_j)}},
\end{equation}
where $i \in \{0,1\}$ means resolution scale as shown in Fig. \ref{fig:network}, $L_{sm}$ is softmax loss, $y_j$ is ground truth label of voxel $j$,  $p_j$ is the predicted possibility, and $w_j  \in  \big\{0,1\big\} $ is the weight of this voxel. The final loss is a summation of all losses as follows:
\begin{equation}
\label{equ:final_loss}
    L =  \sum \limits_i{\alpha_i L_i},
\end{equation}
where $\alpha_i$ is the weight for each scale. We found $\alpha_i = 1$ works well.

\subsection{Implementation Details}
\textbf{Dataset.} We train and evaluate our network on the SUNCG dataset, which is a manually created large-scale synthetic scene dataset \cite{song2016semantic}. It
contains 139368 valid pairs of depth map and complete labels for training, and 470 pairs for testing. Depth maps are converted to volumes with a size of $240\times144\times240$. The ground truth labels are 12-class volumes with 1/4 size of input volume.

\noindent\textbf{Network Details.} The detailed network architecture is illustrated in Fig. \ref{fig:network}. For volumetric data encoding, we use flipped Truncated Signed Distance Function (fTSDF), which can enhance performance because it eliminates strong gradients in empty space \cite{song2016semantic}. The input size of our network is $256^3$, and we put the original fTSDF volume in the middle of input volume. The input volume is down-sampled twice using Max-pooling layer. Then a U-Net architecture follows, which contains six resolution scales, from $64^3$ to $2^3$. Features from encoding stages and decoding stages are summed, and zeros are filled at missing locations. The network uses pre-activation Resnet block in encoding and decoding modules \cite{he2016deep,he2016identity}, and each block has two $3\times3\times3$ convolutions. Down-sampling and up-sampling are implemented by convolution layers with stride 2 and kernel size 2.
SGC is used in resolution scales not less than $32^3$, which account for most of the computation.
Partition operation is performed again once the resolution scale changes, which can help information flow across each other group.

The weight of each voxel is computed by randomly sampling empty and non-empty voxels at a ratio of $1:2$ \cite{song2016semantic}. All non-empty voxels are positive examples. For negative examples, we mainly consider empty voxels around the surface as hard examples, which can be determined by the TSDF value of GT labels ($|TSDF| < 1$). The ratio of hard negative and easy negative examples is 9:1.

\noindent\textbf{Training Policy.} Networks are trained using stochastic gradient descent with a momentum of 0.9. The initial learning rate is 0.1, and L2 weight decay is 1e-4. We train our network for 10 epochs with a batch size of 4, and decay learning rate by a factor of $exp(-0.5)$ in each epoch. In order to reduce training time, we randomly select 40000 samples in each epoch, and the total training time is about 5 days with a GTX TitanX GPU and two Intel E5-2650 CPUs.

\section{Evaluation}
In this section, we evaluate our network on the standard SUNCG test dataset. Both semantic scene completion results and scene completion results are given. Voxel-level IoU evaluation metric is used. Semantic scene completion results are evaluated on both observed and unobserved voxels, and completion results are evaluated on unobserved voxels.
Table \ref{tab:baseline_results} and Table \ref{tab:group-results} show the quantitative results of our network without or with SGC. Fig. \ref{fig:results} shows the qualitative comparison with previous work.
We also give results on real-word noisy NYU dataset \cite{silberman2012indoor}.

\subsection{Comparision to SSCNet}
\begin{figure}
\centering
\includegraphics[width=\textwidth]{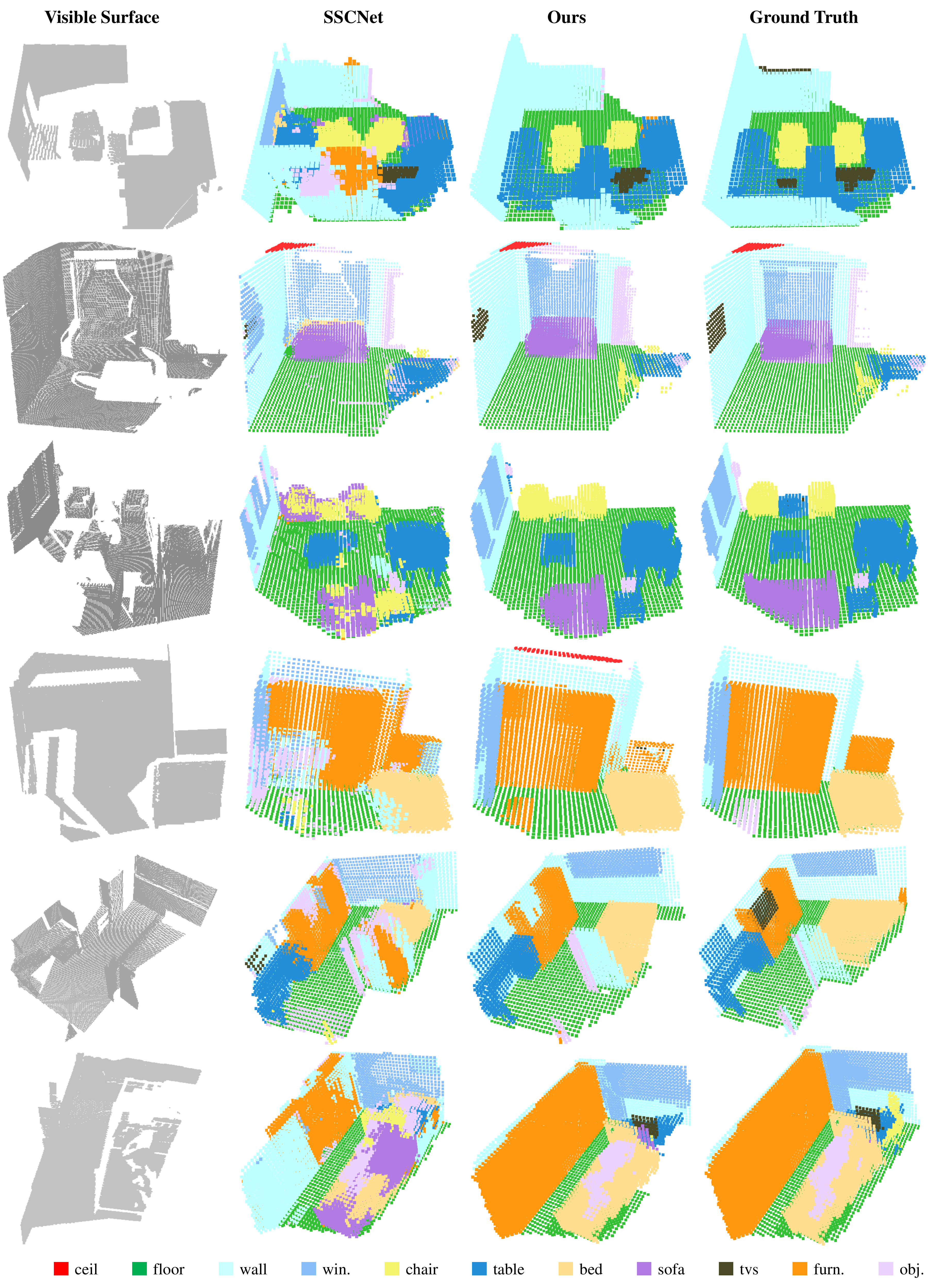}
\caption{Qualitative results of our network and SSCNet. We achieve obviously much better results, such as predictions around object boundaries.}
\label{fig:results}
\end{figure}

Table \ref{tab:baseline_results} shows the result of our baseline network without SGC (group number is 1). We outperform the previous SSCNet by a significant margin, having an improvement of 24.1\% in semantic scene completion and 11.0\% in scene completion, and achieving state-of-the-art results. Our network exceeds SSCNet in almost all classes, especially in small and hard categories such as chair, tvs and objects. We attribute this improvement to the novel SCN architecture that enables the usage of several advanced deep learning techniques such as deeper networks (15-layer vs 57-layer), multiscale network architecture (3 resolution scales vs 8 resolution scales), batch normalization layer \cite{ioffe2015batch} and stacked Resnet style blocks. Fig. \ref{fig:results} shows the visualization results of semantic scene completion from a single depth image. Obviously our baseline network produces visually better results compared to SSCNet, especially around the object boundaries.

\begin{table}[t]
\centering
\caption{Quantitative results of our network and SSCNet on the SUNCG dataset. Scene completion IoU is measured on unobserved voxels, and all non-empty classes are treated as one category. Semantic scene completion IoU is measured on both observed and unobserved voxels. Overall, our method outperforms SSCNet by a large margin. Better results of each category are bold.}
\label{tab:baseline_results}
\resizebox{\textwidth}{!}
{\begin{tabular}{l|ccc|cccccccccccc}
\hline
            & \multicolumn{3}{c|}{scene completion}  & \multicolumn{12}{c}{semantic scene completion}                                                                                                                                        \\ \hline
Method       & prec.         & recall & IoU           & ceil.         & floor & wall          & win.          & chair         & bed           & sofa          & table         & tvs           & furn.         & objs.         & avg.          \\ \hline
SSCNet \cite{song2016semantic}       & 76.3          & \textbf{95.2}   & 73.5          & 96.3          &\textbf{ 84.9}  & 56.8          & 28.2          & 21.3          & 56.0          & 52.7          & 33.7          & 10.9          & 44.3          & 25.4          & 46.4          \\ \hline
Our & \textbf{92.6} & 90.4   & \textbf{84.5} & \textbf{96.6} & 83.7  & \textbf{74.9} & \textbf{59.0} & \textbf{55.1} & \textbf{83.3} & \textbf{78.0} & \textbf{61.5} & \textbf{47.4} & \textbf{73.5} & \textbf{62.9} & \textbf{70.5} \\ \hline
\end{tabular}
}
\end{table}

\subsection{Spatial Group Convolution Evaluation}
This section describes the results of networks with SGC (see Table \ref{tab:group-results}). We conduct experiments on 2,3,4,6 groups with different partition strategies. The efficiency is evaluated with FLOPs, i.e., the number of floating-point multiplication-adds of the whole network. As shown in Table \ref{tab:group-results}, SGC can reduce about $(G-1)/G$ of the whole computation.
Experiments show that 3D sparse CNNs can be sparsity-invariant to some extent, because accuracy only drops about 0.5\% when dividing voxels into two groups even randomly, while only about 50\% voxels preserved in this case. Increasing group number will reduce more computation at the cost of a little drop of performance. Compared to random partition method, fixed pattern partition strategy can give better performance yet requires less computation. For example, 1.7\% IoU enhancement for semantic completion can be achieved using fixed pattern partition method when dividing voxels into four groups.
Overall, SGC can significantly reduce the computation while maintaining accuracy, achieving a drop of only 0.7\% and 1.2\% in terms of IoU for scene completion and semantic completion task while using only 27.8\% computation.

\begin{table}[t]
\centering
\caption{Quantitative IoU (\%) results of networks using SGC with random partition strategy or fixed pattern partition strategy. Both accuracy and FLOPs are given. For fixed pattern partition method, flexible parameters (a,b,c) are also given and we select the best results in our experiments. Best trade-off is bolded.}
\label{tab:group-results}
\resizebox{\textwidth}{!}
{\begin{tabular}{c|c|c|c|l|cl}
\hline
{Group No.} & {Method} & {scene completion} & \multicolumn{2}{c|}{{semantic scene completion}} & \multicolumn{2}{c}{{FLOPs/G}} \\ \hline
1(Baseline)                      &                               & 84.5                                        & \multicolumn{2}{c|}{70.5}                                                 & \multicolumn{2}{c}{79}                             \\ \hline
                                 & Random                        & 83.9                                        & \multicolumn{2}{c|}{69.9}                                                 & \multicolumn{2}{c}{42}                             \\
\multirow{-2}{*}{2}              & Pattern(1,1,1)                & 84.0                                        & \multicolumn{2}{c|}{69.6}                                                 & \multicolumn{2}{c}{39}                             \\ \hline
                                 & Random                        & 82.6                                        & \multicolumn{2}{c|}{67.6}                                                 & \multicolumn{2}{c}{29}                             \\
\multirow{-2}{*}{3}              & Pattern(1,1,1)                & 84.1                                        & \multicolumn{2}{c|}{69.5}                                                 & \multicolumn{2}{c}{27}                             \\ \hline
                                 & Random                        & 83.1                                        & \multicolumn{2}{c|}{67.6}                                                 & \multicolumn{2}{c}{23}                             \\
\multirow{-2}{*}{4}              & Pattern(1,2,3)                & \textbf{83.8}                                        & \multicolumn{2}{c|}{\textbf{69.3}}                                                 & \multicolumn{2}{c}{\textbf{22}}                             \\ \hline
                                 & Random                        & 82.3                                        & \multicolumn{2}{c|}{66.6}                                                 & \multicolumn{2}{c}{17}                             \\
\multirow{-2}{*}{6}              & Pattern(1,2,1)                & 82.6                                        & \multicolumn{2}{c|}{66.9}                                                 & \multicolumn{2}{c}{16}                             \\ \hline
\end{tabular}}
\end{table}

\begin{table}[t]
\centering
\caption{
Influence of SGC on each category. The numbers in third to sixth row mean IoU (\%) drop when using SGC. The best or worst three are underlined or bolded.}
\label{tab:drop}
\resizebox{\textwidth}{!}
{\begin{tabular}{c|c|cccccccccccc}
\hline
{Group No.}               & {Method}   & {  ceil.}     & {floor}      & { wall}       & { win.}       & { chair}         & { bed}        & { sofa} & { table}         & { tvs}       & { furn.}         & { objs.}         & { avg.} \\ \hline
{ Baseline}            & { }        & { 96.6}      & { 83.7}       & { 74.9}       & { 58.9}       & { 55.1}          & { 83.3}       & { 78.0} & { 61.5}          & { 47.4}      & { 73.5}          & { 62.9}          & { 70.5} \\ \hline
{ }                    & { random}  & { {\ul 0.4}} & { {\ul -0.6}} & { {\ul -2.0}} & { -2.9}       & { \textbf{-4.4}} & { -3.0}       & { -3.2} & { -4.1}          & { -3.5}      & { \textbf{-4.4}} & { \textbf{-4.7}} & { -2.9} \\ \cline{2-14}
\multirow{-2}{*}{{ 4}} & { pattern} & { {\ul 0.2}} & { -0.3}       & { -0.8}       & { {\ul 0.7}}  & { \textbf{-3.7}} & { -1.7}       & { -1.5} & { \textbf{-3.1}} & { {\ul 1.1}} & { -1.9}          & { \textbf{-2.1}} & { -1.2} \\ \hline
{ }                    & { random}  & { {\ul 0.1}} & { {\ul -0.7}} & { -3.7}       & { {\ul -2.8}} & { \textbf{-5.7}} & { -2.9}       & { -3.5} & { -5.6}          & { -5.4}      & { \textbf{-6.8}} & { \textbf{-6.0}} & { -3.9} \\ \cline{2-14}
\multirow{-2}{*}{{ 6}} & { pattern} & { {\ul 0.2}} & { {\ul -0.4}} & { -2.3}       & { -3.8}       & { \textbf{-5.5}} & { {\ul -1.8}} & { -3.7} & { \textbf{-6.9}} & { -4.0}      & { -5.5}          & { \textbf{-5.8}} & { -3.6} \\ \hline
\end{tabular}}

\end{table}
Table \ref{tab:drop} shows the detailed semantic scene completion results of different categories using SGC. The accuracies of best and worst three categories compared to baseline network are marked in the table. It can be found that the IoUs of categories with small physical sizes such as chair, furniture, and objects drop more than categories with large size such as ceiling and floor. This may be caused by the fact that those small objects have fewer voxels. Dividing these voxels into different groups may lose important geometric information and makes it harder to distinguish these objects (see chair{'}s leg in Fig. \ref{fig:full_sparse}). While large objects have surplus voxels, sparser voxels can still keep a rough structure. So, a possible future work to increase the accuracy of semantic scene completion is to adaptively handle large easy objects and small hard objects, sampling small objects with high density while sampling large objects with relatively low density.

We also tried sparsity invariant convolution \cite{2017arXiv170806500U} in random partition method, which normalizes convolution by a factor of valid voxels number, but it does not work in our task.

\subsection{Evaluation on NYU dataset}
NYU \cite{silberman2012indoor} contains 1449 depth maps captured by Kinect. Following SSCNet \cite{song2016semantic}, we use Guo et al.{'}s algorithm \cite{guo2015predicting} to generate ground truth annotations for semantic scene completion task. The object categories are mapped based on Handa et al. \cite{handa2016understanding}. We trained the network described above from scrath on NYU dataset. The base of exponential learning rate decay is 0.12 and we trained it for 40 epochs using the whole dataset. Other hyperparameters are same as experiments on SUNCG. Table \ref{tab:nyu_baseline} shows that our network achieves an improvement of 2.0\% in semantic scene completion and 1.1\% in scene completion compared to SSCNet.
Table \ref{tab:nyu_sgc} gives detail results on NYU dataset. It shows that SGC operation is still effective on real data. The fixed pattern partition method gives comparable or even better results than baseline network, and it is consistently better than the random partition method. Note that there exists a gap between the improvements on SUNCG and NYU. We attribute this gap to the fact that misalignment and incomplete annotations are common in the generated labels \cite{guo2015predicting}. This may both mislead the training and evaluation procedures, and it may be unfavorable for our network considering the sparsity geometry representation.

\begin{table}[t]
\centering
\caption{Scene completion (IoU \%) and semantic scene completion results (IoU \%) on NYU dataset.}
\label{tab:nyu_baseline}
\resizebox{\textwidth}{!}
{\begin{tabular}{l|ccc|cccccccccccc}
\hline
            & \multicolumn{3}{c|}{scene completion}  & \multicolumn{12}{c}{semantic scene completion}                                                                                                                                        \\ \hline
Method       & prec.         & recall & IoU           & ceil.         & floor & wall          & win.          & chair         & bed           & sofa          & table         & tvs           & furn.         & objs.         & avg.          \\ \hline
SSCNet & 57.0        & \textbf{94.5}        & 55.1      & 15.1  & \textbf{94.7}  & 24.4 & 0    & 12.6  & 32.1 & 35   & \textbf{13}    & \textbf{7.8} & 27.1  & 10.1  & 24.7 \\ \hline
Ours       & \textbf{71.9}        & 71.9        & \textbf{56.2}      & \textbf{17.5}    & 75.4  & \textbf{25.8} & \textbf{6.7}    & \textbf{15.3}  & \textbf{53.8} & \textbf{42.4} & 11.2     & 0   & \textbf{33.4}  & \textbf{11.8}  & \textbf{26.7} \\ \hline
\end{tabular}
}
\end{table}

\begin{table}[t]
\centering
\caption{Results (IoU\%) of networks with SGC using different partition strategies on NYU dataset. (SSC stands for semantic scene completion)}
\label{tab:nyu_sgc}

\begin{tabular}{c|c|c|c|c|c|c|c}
\hline
Group No.    & 1        & \multicolumn{2}{c|}{2} & \multicolumn{2}{c|}{3} & \multicolumn{2}{c}{4} \\ \hline
Method       & baseline & random & pattern & random & pattern & random & pattern \\ \hline
SSC & 26.4     & 24.1   & 26.5          & 23     & \textbf{26.7} & 22.6   & 25.9          \\ \hline
Completion   & 55.7     & 53     & 54.8          & 52.2   & \textbf{56.2} & 52.6   & 55.1          \\ \hline
\end{tabular}

\end{table}
\begin{figure}[t]
\centering
\includegraphics[width=\textwidth]{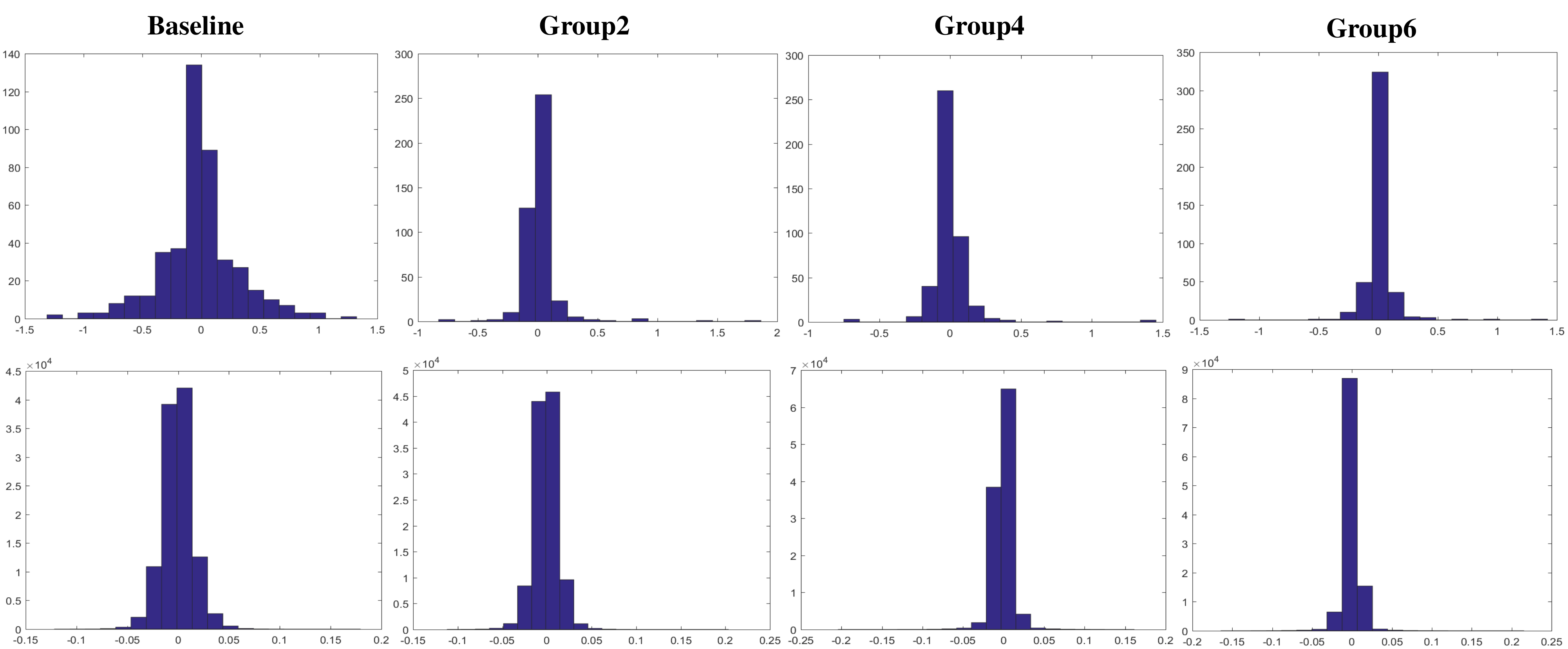}
\caption{Histograms of learned weight values of SCN and SGC with different groups. The first row shows the statistics of the first convolution layer, and the second row shows that of the last convolution layer. Filters of SGC have ``sharper" histograms.}
\label{fig:weight_visual}
\end{figure}

\section{Discussion}
\subsection{What does Spatial Group Convolution learn?}
In Fig. \ref{fig:weight_visual}, we visualize the histograms of learned weight values of networks without or with SGC using random partition method. It can be observed that filters of SGC have ``sharper" histograms while normal SCN filters have relative ``flat" histograms, which means the values of SGC filters are pretty close. The histograms become ``sharper" with the increase of group number. This may be caused by that filters of SGC need to be invariant to different sparsity patterns, so the values of filters at different locations had better be close to adapt to different sparsity patterns.

\begin{figure}[t]
\centering
\includegraphics[height=5cm]{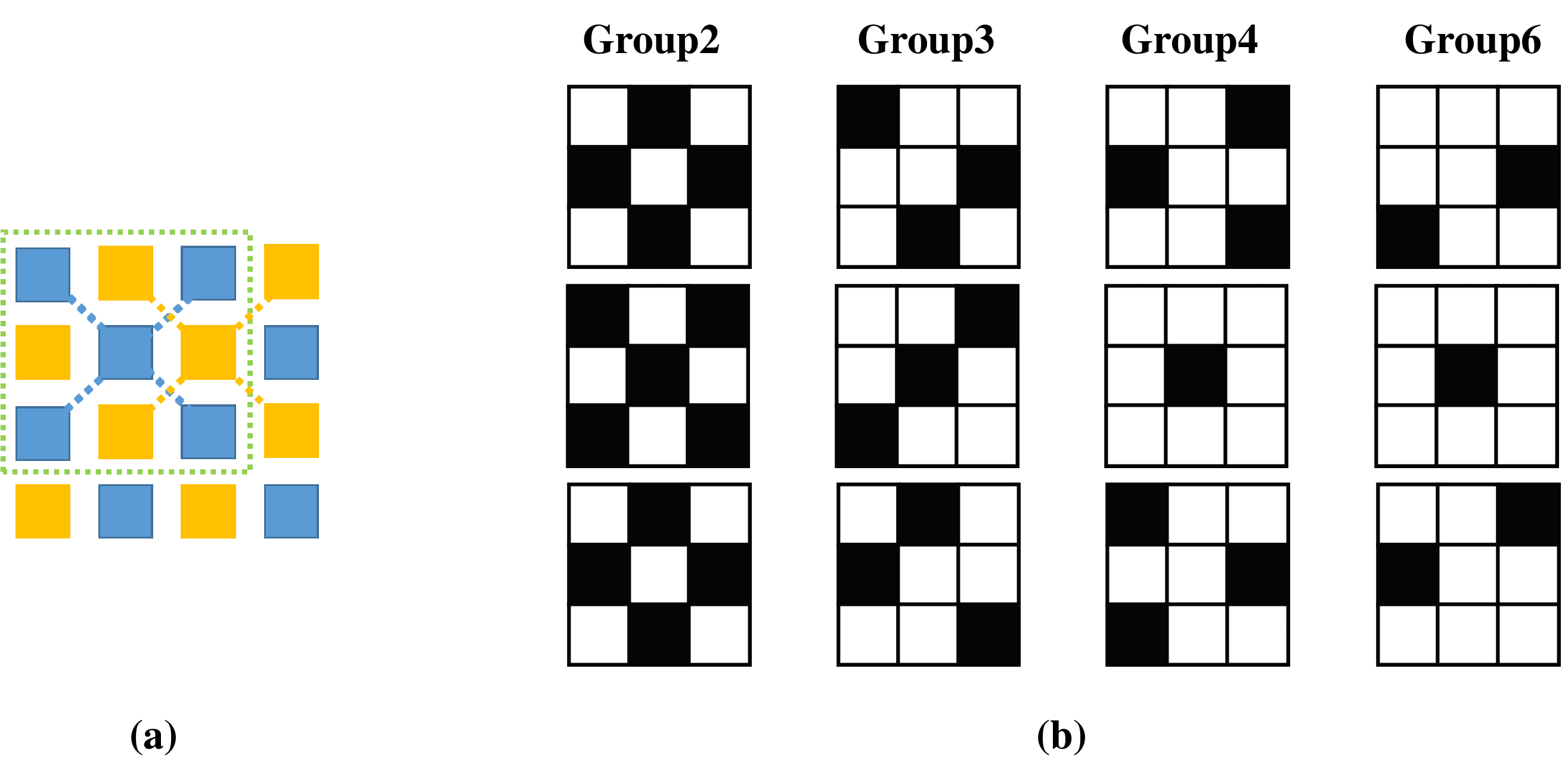}
\caption{Illustration of SGC with fixed pattern partition. (a) shows that for a $3\times3$ kernel, an ``X" shape filter is learned when partitioning voxels into two groups. (b) shows the learned $3\times3\times3$ filters in Table \ref{tab:group-results}. Filters are drawn by slice.}
\label{fig:pattern}
\end{figure}

As for SGC with fixed pattern partition, we find it learned an irregular convolutional kernel. In Fig. \ref{fig:pattern}a, we show a simple case in a 2D convolution which divides voxels into two groups. The valid convolutional kernel shape is always ``X" because the sparsity pattern keeps the same when sliding the convolutional kernel. Fig. \ref{fig:pattern}b shows the valid convolutional filter shapes used in Table \ref{tab:group-results}.

\subsection{Information flow among different groups}

The SGC operation partitions voxels into different groups. During convolution, different groups are independent and have no information flow across each other. However, after SGC, the voxels are gathered and fed into down-sampling convolution or up-sampling deconvolution layers, in which information of different groups can communicate. Besides, we also explored more complicated methods to help information exchange among different groups. For example, Shuffled SGC, which is inspired by ShuffleNet \cite{zhang2018shufflenet}. ShuffleNet uses channel shuffle to help information flow across feature channels, while we shuffle the features across spatial dimensions which is implemented by using different partition patterns in the two convolution layers of Resnet block. But no obvious improvement is observed in our case.


\section{Conclusions}
The paper presents an efficient semantic scene completion network with Spatial Group Convolution. SGC partitions feature maps into different groups along the spatial dimensions and can significantly reduce the computation with slight loss of accuracy. Besides, we propose a novel end-to-end sparse convolutional network architecture for 3D semantic scene completion and set a new accurancy record on the SUNCG dataset.

\section*{Acknowledgment}
This work was supported in part by National Key Research and Development Program of China (2017YFC0108000), National Natural Science Foundation of China (81427803, 81771940, 61132007, 61172125, 61601021, and U1533132), Beijing Municipal Natural Science Foundation (7172122,L172003), and Soochow-Tsinghua Innovation Project (2016SZ0206).

\bibliographystyle{splncs04}
\bibliography{jiahui-eccv}




\end{document}